\documentclass[10pt,twocolumn,letterpaper]{article}

\usepackage{cvpr}              
\usepackage{setspace}
\usepackage{amsmath}
\usepackage{amssymb}
\usepackage{xcolor}
\usepackage{booktabs}
\usepackage[table]{xcolor}
\usepackage{caption}
\usepackage{subcaption}
\usepackage{tabularx}
\usepackage{bm} 
\usepackage[ruled]{algorithm2e}
\usepackage{setspace}
\usepackage{graphicx}

\newcommand{\GrayComment}[1]{%
  \tcp*{\textcolor{gray}{\footnotesize{$\triangleright$~#1}}}%
}



\definecolor{cvprblue}{rgb}{0.21,0.49,0.74}
\definecolor{codegreen}{rgb}{0,0.6,0}
\definecolor{codegray}{rgb}{0.5,0.5,0.5}
\definecolor{codepurple}{rgb}{0.58,0,0.82}
\definecolor{backcolour}{rgb}{0.95,0.95,0.92}
\definecolor{mydarkred}{rgb}{0.8,0.02,0.02}
\definecolor{mydarkorange}{rgb}{0.40,0.2,0.02}
\definecolor{mypurple}{RGB}{111,0,255}
\definecolor{myred}{rgb}{1.0,0.0,0.0}
\definecolor{mygold}{rgb}{0.75,0.6,0.12}
\definecolor{mydarkgray}{rgb}{0.66, 0.66, 0.66}
\definecolor{lightblue}{rgb}{0.93,0.95,1.0}
\definecolor{mygray}{gray}{0.9}
\definecolor{history}{RGB}{255,110,1}
\definecolor{start}{RGB}{255,129,166}
\usepackage[pagebackref,breaklinks,colorlinks,allcolors=cvprblue]{hyperref}

\def\paperID{1926} 
\def\confName{CVPR}
\def\confYear{2026}

\title{Markovian Scale Prediction: A New Era of Visual Autoregressive Generation}

\author{
Yu Zhang$^1$\quad 
Jingyi Liu$^1$$^\dagger$\quad 
Yiwei Shi$^2$\quad 
Qi Zhang$^1$\quad 
Duoqian Miao$^1$$^*$ \\
Changwei Wang$^1$\quad 
Longbing Cao$^3$ \vspace{0.2cm}\\
$^1$Tongji University \quad 
$^2$University of Bristol \quad 
$^3$Macquarie University \\
\vspace{0.3cm}\\
Project Page: \href{https://luokairo.github.io/markov-var-page/}{https://luokairo.github.io/markov-var-page/}
}

\begin{document}
\twocolumn[{

\maketitle
\begin{center}
\vspace{-0.5cm}
    \captionsetup{type=figure}
    \includegraphics[width=\linewidth]{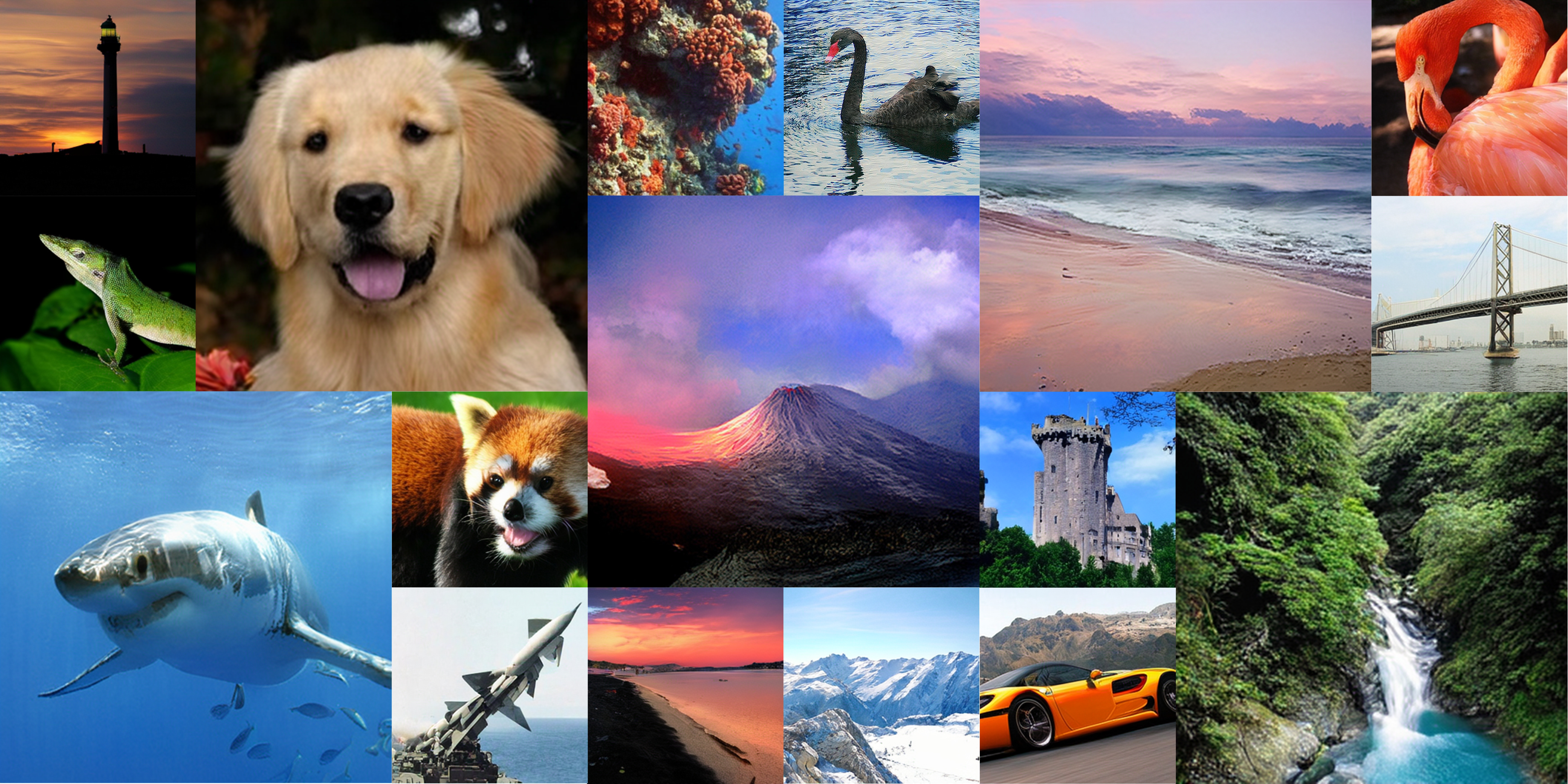}
    \captionof{figure}{Visualization of generated images from our Markov-VAR at 256×256 or 512×512 on ImageNet benchmark.}
    \vspace{0.3cm}
    \label{teaser}
\end{center}
}]

\begin{abstract}
\makeatletter
\renewcommand*{\@makefnmark}{}
\footnotetext{
$^\dagger$ Project Lead, $^*$ Corresponding Author.
}
Visual AutoRegressive modeling (VAR) based on next-scale prediction has revitalized autoregressive visual generation. Although its full-context dependency, i.e., modeling all previous scales for next-scale prediction, facilitates more stable and comprehensive representation learning by leveraging complete information flow, the resulting computational inefficiency and substantial overhead severely hinder VAR's practicality and scalability. This motivates us to develop a new VAR model with better performance and efficiency without full-context dependency. To address this, we reformulate VAR as a non-full-context Markov process, proposing Markov-VAR. It is achieved via Markovian Scale Prediction: we treat each scale as a Markov state and introduce a sliding window that compresses certain previous scales into a compact history vector to compensate for historical information loss owing to non-full-context dependency. Integrating the history vector with the Markov state yields a representative dynamic state that evolves under a Markov process.  Extensive experiments demonstrate that Markov-VAR is extremely simple yet highly effective: Compared to VAR on ImageNet, Markov-VAR reduces FID by 10.5\% (256×256) and decreases peak memory consumption by 83.8\% (1024×1024). We believe that Markov-VAR can serve as a foundation for future research on visual autoregressive generation and other downstream tasks.
\end{abstract}   
\newpage

\twocolumn[{
\begin{center}
    \captionsetup{type=figure}
    \includegraphics[width=\linewidth]{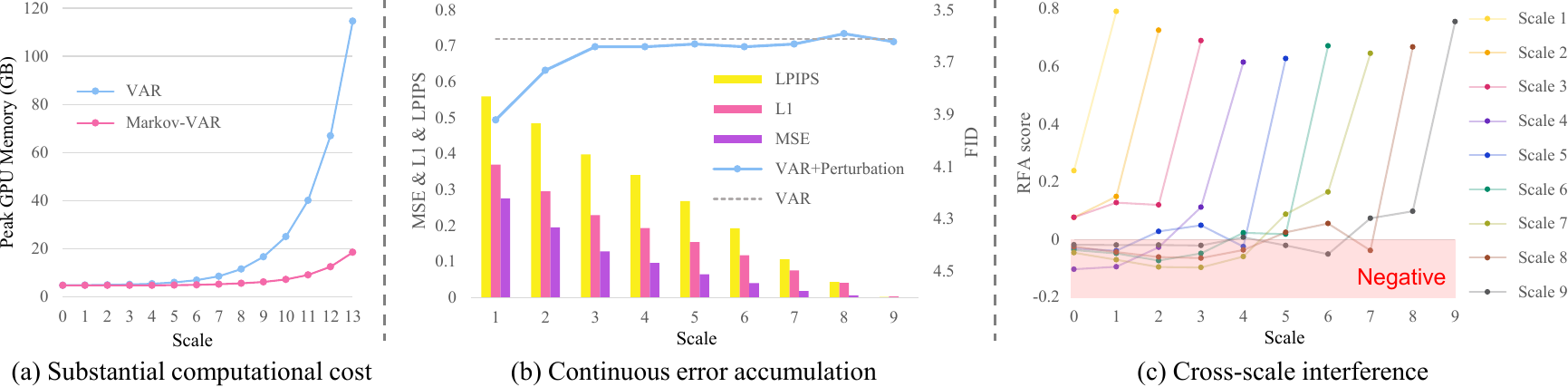}
    \captionof{figure}{Observations of the challenges caused by full-context dependency. (a) Comparison of peak computation state (Activations + KV Cache) memory consumption between depth-24 VAR and Markov-VAR on generating 1024×1024 images with a batch size of 25. (b) Metrics and FID performance of VAR under perturbations injected at different scales. MSE, L1 and LPIPS jointly decrease as the perturbation injection scale shifts larger, indicating that early injected perturbations cause greater performance degradation. This is also evidenced by VAR's largest FID drop at the first injection scale. (c) Residual-Feature Alignment scores (RFA) between each scale and its every previous scale. It is calculated as the cosine similarity between the output residual feature of the current scale and each input feature of all previous scales, combined with $1\times1$ convolution projection and square root operation, and preserves the directional contribution.}
    \label{observation}
\end{center}
}]

\section{Introduction}
As a general modeling paradigm, autoregressive modeling has dominated text generation ~\cite{wan2023efficient,wan2024d2o,touvron2023llama,chung2024scaling,achiam2023gpt, lu2025easytext, shi2024fonts, shi2025wordcon} and multimodal understanding~\cite{xie2025collaboration,Chen_2024_CVPR,li2024llava}. Yet, due to suboptimal generation quality, it was not widely adopted in visual generation tasks for a long time. Recently, Visual AutoRegressive modeling (VAR)~\cite{VAR} converts next-token prediction to next-scale prediction, further unleashing the potential of the autoregressive modeling to generate high-quality visual content, making autoregression once again a mainstream modeling paradigm for visual generation~\cite{kumbong2025hmar,shao2025continuous,ren2024mvardecoupledscalewiseautoregressive, han2025infinity, hui2025autoregressive, zhang2025easycontrol}. However, VAR's full-context dependency, requiring attention over all previous scales for next-scale prediction, can introduce certain challenges:

\textit{\textbf{I. Substantial computational cost.}} As the scale increases, the number of tokens grows quadratically. Moreover, cumulative modeling across multiple previous scales in VAR accelerates the superlinear increase in computational cost, as shown in Figure~\ref{observation} (a). Such substantial computational cost not only slows down training and inference, but also severely limits VAR's scalability and practicality.

\textit{\textbf{II. Continuous error accumulation.}} As a chain-based modeling paradigm, autoregression's unidirectional causality chain prevents early prediction errors from being corrected and leads to their continuous propagation. Figure~\ref{observation} (b) shows that early perturbations impact the performance more severely than late ones, indicating that errors continuously accumulate during error propagation. In addition, VAR exacerbates the error accumulation in two aspects. First, its full-context dependency repeatedly utilizes and iteratively accumulates errors from previous scales. Second, rapid growth of the token sequence length extends the chain of error propagation and accumulation. This cross-scale error accumulation significantly degrades both the quality and stability, especially for high-resolution visual generation. 

\label{ana:interference.}
\textit{\textbf{III. Cross-scale interference.}} VAR's coarse-to-fine modeling requires learning distinctive scale-specific representations at each scale. However, for the current scale, full-context dependency leads attention to aggregate all previous scales, where mixed information of multiple scales makes attention and gradients from different scales compete or conflict with each other in the shared feature space, thereby suppressing the learning of distinctive representations at the current scale and limiting the improvement of generation quality. As illustrated in Figure \ref{observation} (c), we compute RFA scores to quantify the directional contribution of the input feature from previous scales to the current predicted residual feature, further measuring cross-scale consistency in learning the current representation. We find that early scales typically have a negative impact on learning distinctive representations at the current scale.

These challenges motivate us to explore \textit{whether it is feasible to develop a visual autoregressive generation model without full-context dependency that outperforms traditional VAR in both performance and efficiency}. Inspired by the concept of sufficient statistics~\cite{tishby2000informationbottleneckmethod,7133169,alemi2019deepvariationalinformationbottleneck} in information theory, we suppose that each node in continuous chain-based propagation inherently maintains historical information to a certain extent. Appropriately distilling this into a representative dynamic state can achieve effective prediction without relying on all original historical information.

In light of this motivation, we develop Markov-VAR, a visual autoregressive generation model with Markov process. Markov-VAR refines next-scale prediction into \textit{\textbf{Markovian scale prediction}}, where each scale prediction depends only on the current one rather than all previous scales. Chain-based unidirectional autoregressive modeling makes the current scale already encoding representative historical information for feature prediction~\cite{wiewiora2005learning,yang2022discrete,mao2020information}. Thus, we can treat each scale as a Markov state. However, non-full-context dependency inevitably discards substantial original historical information compared to full-context dependency. We establish a lightweight history compensation mechanism to enrich the Markov state for historical information. Specifically, we employ a sliding window that compresses certain previous scales within it into a compact history vector, which is then integrated with the Markov state to construct a representative dynamic state. By modeling these dynamic states as a Markov process, Markov-VAR achieves Markovian scale prediction.

Extensive experiments demonstrate Markov-VAR's superior performance and efficiency. For image generation on ImageNet~\cite{deng2009imagenet}, compared to VAR, Markov-VAR reduces FID from 3.61 to 3.23 (256×256) and remarkably decreases peak memory consumption of the computation state from 117.9GB to 19.1GB (1024×1024). As a foundation model, its performance and efficiency may become more promising when combined with other enhancement or acceleration techniques. Moreover, Markov-VAR outperforms many alternative-paradigm models or improved VAR-like variants in various comparisons. Our contributions are as follows:

\vspace{0cm}
\begin{itemize}
    \vspace{0cm}
    \item We advance visual autoregressive generation by exploring modeling without full-context dependency, refining next-scale prediction as Markovian scale prediction to reformulate VAR as a non-full-context Markov process.
    \vspace{0cm}
    \item We propose Markov-VAR with a history compensation mechanism to mitigate the historical information loss in the Markov process. Markov-VAR comprehensively surpasses VAR in both performance and efficiency, extremely simple yet highly effective.
    \vspace{0cm}
    \item We publicly release the full series of Markov-VAR model weights, hoping to serve as a foundation model and facilitate future research on various tasks.
\end{itemize}

\section{Related Work}
\label{sec:rel}
\paragraph{Visual Autoregressive Generation.} 
Early visual autoregressive generation models such as PixelRNN~\cite{van2016pixel}, PixelCNN~\cite{van2016conditional} and PixelCNN++~\cite{ salimans2017pixelcnn++} achieved pixel-level conditional modeling, yet this paradigm suffers from limited long dependency and extremely slow sampling. Subsequent works~\cite{ramesh2021zero,esser2021imagebart,van2017neural,zhu2024scaling,ding2021cogviewmasteringtexttoimagegeneration,rqvae,yu2022scalingautoregressivemodelscontentrich,song2025makeanything,wan2024grid} treating image synthesis as a token prediction task relied on Vector Quantization (VQ)~\cite{van2017neural} to discretize continuous feature maps and autoregressively modeling visual token sequence. However, their raster-scan prediction order conflicts with the spatial structure of the images, leading to inferior performance compared to other alternative-paradigm models~\cite{kumari2023multi,guo2024i2v,ddim,ddpm, ho2020denoising,shen2025efficient,goodfellow2014generative, chen2025transanimate}. Recently, VAR~\cite{VAR} reformulates next-token prediction as next-scale prediction, further improving the generation quality of visual content. Some VAR-like variants, such as Infinity~\cite{han2025infinity}, HART~\cite{tang2024hart} and VAR-CLIP~\cite{zhang2024var} extend VAR to text-to-image generation. Other further improved VAR-like models~\cite{li2024controlvar, guo2025fastvar, rajagopalan2025restorevar, xie2024litevar,wu2025nestedautoregressivemodels,jiao2025flexvarflexiblevisualautoregressive,InfinityStar,zhuang2025vargpt} perform various tasks, such as image super-resolution~\cite{qu2025visual}, 3D Object Generation~\cite{chen2025sar3d} and image editing~\cite{wang2025training}.

\paragraph{Chain-based Markov Modeling.} The Markov assumption~\cite{markov1906rasprostranenie} that posits that the current state depends only on its immediate predecessor, has been widely employed for research in deep learning. Diffusion-based models~\cite{croitoru2023diffusion, zhang2025enhancing,zhu2025llada,nie2025large} have achieved remarkable success in various tasks based on this principle. Chain-based Markov modeling has been introduced to alleviate efficiency challenges of long-context modeling. MARCOS~\cite{liu2025marcos} and MCoT~\cite{yang2024markov} introduce structured memory nodes that reformulate chain-of-thought reasoning into a Markovian process for more lightweight inference. In visual autoregressive modeling, although HMAR~\cite{kumbong2025hmar} pioneeringly introduces Markov dependency to improve generation performance, it comes at the cost of increasing inference steps and the token sequence length. MVAR~\cite{zhang2025mvar} innovatively introduces spatial-Markov attention, focusing on reduce computational complexity. We argue that historical information preservation, more concise modeling, and more comprehensive scaling validation are essential for further advancing Markov visual autoregressive modeling. Therefore, we revisit these challenging trade-off problems and employ chain-based Markov modeling to develop a visual autoregressive foundation model that reconciles performance with efficiency.

\section{Method}

\subsection{Preliminary: Next-Scale Prediction}
VAR~\cite{VAR} reformulates autoregressive modeling from next-token prediction to next-scale residual feature prediction, which generates images in a coarse-to-fine manner based on full-context dependency. The modeling process involves $T$ multi-scale residual features $\{R_1, R_2,\cdots,R_T\}$ defined by a size set $\{S_1 \times S_1, S_2 \times S_2, \cdots, S_T\times S_T\}$. At the $t$-th scale, VAR predicts the residual feature $R_t \in \mathbb{R}^{S_t \times S_t}$ based on all previous residual scales $R_{<t}=\{R_1, R_2,\cdots,R_{t-1}\}$. The autoregressive likelihood is formulated as follow:
\begin{equation}
    p(R_1, R_2,\cdots,R_T)
    ~=~
    \prod_{t=1}^{T}
    p(R_t ~\bigl|~\langle\mathrm{sos}\rangle, R_{<t}),
    \label{eq:var}
\end{equation}
where $R_{<t}$ denotes the ``prefix" of $R_t$, and $\langle\mathrm{sos}\rangle$ is the start conditional embedding.

\begin{figure*}[t]
    \centering
    \includegraphics[width=\linewidth]{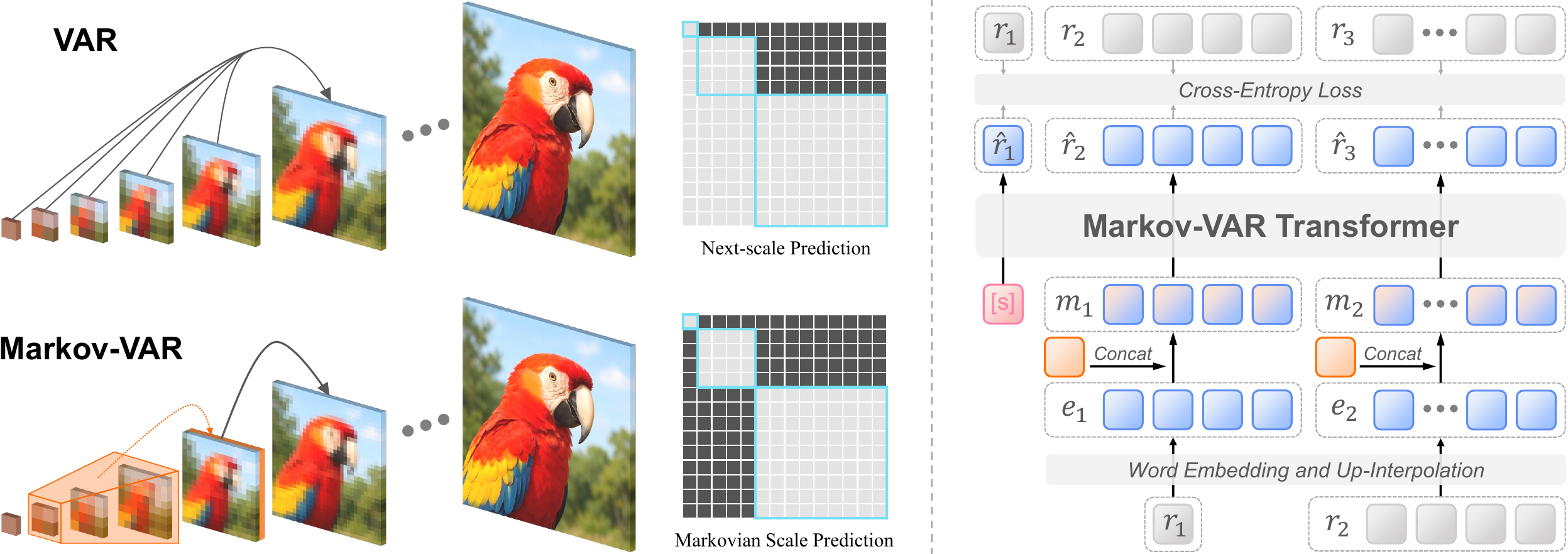}
    \caption{Left: Comparison of modeling process between VAR and Markov-VAR when predicting the 6-th scale \& Comparison of visual context between next-scale prediction and Markovian scale prediction during generation. Markov-VAR utilizes a \textcolor{history}{history compensation mechanism} to enrich the current scale for historical information. Right: The overall framework of Markovian scale prediction with Markov-VAR Transformer. \textcolor{start}{[S]} is the start token with condition embedding.}
    \label{framework}
\end{figure*}

\subsection{Markovian Scale Prediction}
\label{sub:msp}
In contrast to VAR, Markov-VAR is based on Markovian scale prediction. The prediction process treats each scale as a Markov state and models it as a chain-based Markov process. The size set $\{S_1 \times S_1, S_2 \times S_2, \cdots, S_T\times S_T\}$ defines $T$ multi-scale residual features $\{R_1, R_2,\cdots,R_T\}$. At the $t$-th scale, Markov-VAR predicts the Markov state $M_t \in \mathbb{R}^{S_t \times S_t}$ based on the Markov state $M_{t-1}$ (the current residual feature $R_{t-1}$). Therefore, Markovian autoregressive likelihood can be formulated as follows:   

\begin{equation}
    p(R_1, R_2,\cdots,R_T)
    ~=~
    \prod_{t=1}^{T}
    p(R_t ~\bigl| M_{t-1}),
    \label{eq:var2}
\end{equation}
where Markovian state $M_t=f_{\phi}(R_t, M_{t-1})$ is a representative dynamic state for the residual feature $R_t$, and $f_{\phi}(~)$ is a state update function, $M_0=\langle\mathrm{sos}\rangle$. 

\subsection{Markov-VAR}
\label{method}
Figure~\ref{framework} illustrates the comparison between VAR and Markov-VAR, and more details of Markov-VAR.

\paragraph{Markov State.} 
Classical information theory~\cite{tishby2000informationbottleneckmethod} and its applications in deep learning~\cite{7133169,alemi2019deepvariationalinformationbottleneck} indicate that: information $I(c_{<t}; c_{t})$ of the complete history $c_{<t}$ with the current time $c_{t}$ is highly redundant. There exists a sufficient statistic $c_{t-1}$ such that: $I(c_{t-1}; c_{t}) = I(c_{<t}; c_{t})$. Therefore, we argue that continuous chain-based modeling can propagate this sufficient statistic, enabling effective prediction of subsequent scales only based on the current scale. This theory allows us to naturally treat each scale as a Markov state. 

\paragraph{History Compensation Mechanism.} While each scale can be directly treated as a Markov state and modeled in a Markov process,  it must be acknowledged that Markovian scale prediction performs suboptimally compared to next-scale prediction based on full-context dependency when the current scale cannot possess the complete sufficient statistic or substantial historical information is discarded. To maintain the Markov process while further improving prediction performance, we design a history compensation mechanism to supplement historical information for prediction.

We set a sliding window $\mathcal{W}$ of size $N$ to store the previous $N$ continuous scales. For the $t$-th residual scale $R_t$, the corresponding sliding window $\mathcal{W}_t$ is denoted as:
\begin{equation}
    \mathcal{W}_t=\{E_{t-1}, E_{t-2}, \dots, E_{t-N}\},
\end{equation}
where $E_{t}\in \mathbb{R}^{n^{t} \times d}$ denotes the embedded feature at the $t$-th scale, which is obtained through word embedding and up-interpolation of the residual scale $R_{t-1}$. $n^t$ is the number of tokens at the $t$-th scale and $d$ is the dimension of token.

We define the token sequence of $E_t$ as $X_t\in \mathbb{R}^{n^t \times d}$. Thus, the overall token sequence $\hat{X}_t$ of all scales within the sliding window is obtained by a concat operation: $\hat{X}_t=Concat(X_{t-1},X_{t-2},\dots,X_{t-N})$. We aggregate $\hat{X}_t$ into a fixed-dimensional history vector $h_t$ via cross-attention: 
\begin{equation}
    h_{t-1}=\text{Attn}(q,\hat{X}_t,\hat{X}_t),
\end{equation}
where $q$ is a learnable query representing the global state, and $\hat{X}_t$ provides the key–value pairs. 

When predicting the $t$-th residual scale $R_t$, we broadcast the history vector $h_{t-1}$: $H_{t-1} = \mathbf{1}_{n_{t-1}} h_{t-1}^{\mathsf{T}}$, and concatenate history vectors $H_{t-1}$ and feature scale $E_{t-1}$ element-wise to obtain a representative dynamic state $M_{t-1}$:
\begin{equation}
M_{t-1}=Concat(E_{t-1},H_{t-1}).
\end{equation}

After modeling the evolution of representative dynamic states as a Markov process, Markov-VAR starts to operate.

\subsection{Training Strategy}

Following the conventional autoregressive paradigm, we construct the ground-truth residual feature sequence $R$ as the supervision target for Markovian scale prediction. 

Markov-VAR is trained under the teacher-forcing scheme while further adopting the Markovian attention shown in Figure~\ref{framework}, restricting each scale to attend only to its current state. Other main strategies remain consistent with standard visual autoregressive training. 

Algorithm~\ref{alg:training} shows the training process, where modules with subscript $\theta$ are learnable. $\hat{f}$ denotes the final target feature of Markov-VAR. $\mathcal{F} = [\langle\mathrm{sos}\rangle, X_1, \cdots, X_{T-1}]$ represents a queue of token sequences. The overall scale sequence $\hat{R}$ consist of the residual features predicted for each scale. Finally, the training loss $\mathcal{L}$ is calculated by the cross-entropy (CE)~\cite{shannon1948mathematical} between the ground-truth $R$ with $\hat{R}$.

\begin{algorithm}[t]
\setstretch{1.0}
\caption{Training Process of Markov-VAR}
\label{alg:training}
\SetKwData{GT}{$[R_1, \dots, R_T]$}
\SetKwData{Win}{$\mathcal{W}$}
\SetKwData{Hist}{$h_t$}
\SetKwData{Finput}{$\mathcal{F}$}
\SetKwData{State}{$M_{t-1}$}
\SetKwData{Pred}{$\hat{R}_t$}
\KwIn{\GT, $N$, $(S_t \times S_t)_{t=1}^{T}$}
\KwOut{Trained model parameters $\theta$}
$M = [M_0, \cdots, M_{T-1}]$,~~~ $M_0 \gets \langle\mathrm{sos}\rangle$ ;\\  
\Finput $\gets$ \textit{Queue}();~~~  
\Win $\gets$ \textit{Queue}(); \\ 
$Queue\_Push(\Finput, \langle\mathrm{sos}\rangle)$; ~~~$\hat{f} \gets tensor(0)$ \;
\ForAll{$t = 1$ \KwTo $T-1$}{
    $\hat{f} = \hat{f} +Up(R_t, S_T)$ \GrayComment{Add residual}
    $E_t = Embed_{\theta}(Down(\hat{f}, S_{t+1}))$ \;
    $Queue\_Push(\Finput, E_t)$ 
}

\ForAll{$t = 1$ \KwTo $T-1$}{
    \If{$\Win.size() == N$}{
        $Queue\_Pop(\Win)$ \GrayComment{Slide window}
    }
    $X_t=Queue\_Front(\Finput)$ \;
    $Queue\_Pop(\Finput)$ \;
    $Queue\_Push(\Win, X_t)$ \GrayComment{Slide window}
    $\hat{X}_t = Concat(\Win)$ \;
    $H_t = Broadcast(Attn(q_{\theta},\hat{X}_t, \hat{X}_t))$ \;
    $M_t = Concat(\Finput.Front(), H_t)$ \GrayComment{Update}
}
$[\hat{R}_1, \cdots, \hat{R}_t] \gets Model_{\theta}([M_0, \cdots,M_{T-1}])$ \;
$\mathcal{L} = \sum_{t=1}^{T} CE(\hat{R}_t, R_t)$ \GrayComment{Compute loss}
$\mathcal{L}.Backward()$ \GrayComment{Update $\theta$ via $\mathcal{L}$}
\Return{$\theta$}
\end{algorithm}
\vspace{-0.2cm}

\section{Experiment}
\subsection{Experiment Settings}
\paragraph{Models and Baselines.} We construct Markov-VAR models with depths $d$ of 16, 20, and 24 layers, the network structure is set as: the width $w = 64d$, the number of attention heads $h=d$ and the dropout rate $dr=0.1\cdot d/24$. We utilize Rotary Positional Embedding~\cite{su2023roformerenhancedtransformerrotary} for learnable positional encoding, and adopt LLaMA-style~\cite{touvron2023llama} attention and MLP blocks  following~\cite{tang2024hart}. For image tokenization, we employ the pre-trained multi-scale VQ-VAE tokenizer from VAR. To evaluate the effectiveness of our design, we compare Marokov-VAR with the VAR~\cite{VAR} and other VAR-like models on comparable model sizes. We further include comparisons with broad generative models of alternative paradigms, such as diffusion models, GANs, and next-token prediction autoregressive models, to provide a comprehensive evaluation.

\vspace{-0.2cm}
\paragraph{Datasets and Metrics.} We conduct the pre-training of Markov-VAR on the ImageNet-1K dataset~\cite{deng2009imagenet}, following the standard class-to-image generation model setting. For quantitative evaluation, we use Fréchet Inception Distance (FID), Inception Score (IS), Precision, and Recall as the primary metrics to comprehensively assess generation quality. We evaluate efficiency in terms of inference time and peak GPU memory consumption of computation state, which is mainly composed of Transformer activations and KV cache.
\vspace{-0.2cm}
\paragraph{Implementation Details.} All models are trained in similar settings, with a base learning rate of $8\times10^{-5}$, using the AdamW optimizer~\cite{loshchilov2017decoupled} with $\beta_1=0.9$ and $\beta_2=0.95$.
The batch size ranges from 768 to 1536, and the training epochs vary from 200 to 400, depending on the model depth. Training is implemented on 8 NVIDIA H200 GPUs. Evaluation is conducted on a single NVIDIA H200 GPU.

\begin{table}[h]
\vspace{0cm}
\centering
\small
\caption{Performance comparison on ImageNet 256×256 class-conditional generation with VAR and VAR-like models. Evaluation metrics include Fréchet Inception Distance (FID) and Inception Score (IS). Precision and Recall jointly assess the fidelity–diversity trade-off of generated images. `$\downarrow$' and `$\uparrow$' indicate that lower or higher values are preferable.}
\vspace{-0.2cm}
\label{tab:var_like}
\resizebox{0.48\textwidth}{!}{
\begin{tabular}{c|c|cc|cc}
\toprule
\textbf{Model} & \textbf{Param} & \textbf{FID}$\downarrow$ & \textbf{IS}$\uparrow$ & \textbf{Precision}$\uparrow$ & \textbf{Recall}$\uparrow$ \\
\midrule
M-VAR-\textit{d20}~\cite{ren2024mvardecoupledscalewiseautoregressive} & 900M & 2.41 & 308.4 & 0.85 & 0.58 \\
FlexVAR-\textit{d24}~\cite{jiao2025flexvarflexiblevisualautoregressive} & 1.0B & 2.21 & 299.1 & 0.83 & 0.59 \\
HMAR-\textit{d24}~\cite{kumbong2025hmar} & 1.3B & 2.10 & 324.3 & 0.83 & 0.60 \\
NestAR-H~\cite{wu2025nestedautoregressivemodels} & 1.3B & 2.22 & 342.4 & 0.79 & 0.57 \\
VAR-\textit{d16}\,\cite{VAR} & 310M & 3.61 & 225.6 & 0.81 & 0.52 \\
VAR-\textit{d20}\,\cite{VAR} & 600M & 2.67 & 254.4 & 0.81 & 0.57 \\
VAR-\textit{d24}\,\cite{VAR} & 1.0B & 2.17 & 271.9 & 0.81 & 0.59 \\
VAR-\textit{d30}\,\cite{VAR} & 2.0B & 2.14 & 275.4 & 0.80 & 0.60 \\
\rowcolor{lightblue}
Markov-VAR-\textit{d16} & 329.0M & 3.23 & 256.2 & 0.84 & 0.52 \\
\rowcolor{lightblue}
Markov-VAR-\textit{d20} & 623.2M & 2.44 & 286.1 & 0.83 & 0.56 \\
\rowcolor{lightblue}
Markov-VAR-\textit{d24} & 1.02B & 2.15 & 310.9 & 0.83 & 0.59 \\
\bottomrule
\end{tabular}
}
\end{table}

\begin{table*}[htb]
\centering
\small
\caption{Board performance comparison on class-to-image generation on ImageNet $256\times256$ benchmark with more alternative-paradigm models. Step metric accounts for the number of sampling steps during inference. The suffix `-re' denotes rejection sampling.}
\vspace{-0.2cm}
\label{tab:board}

\begin{tabular}{c|c|c|cc|cc|c}
\toprule
\textbf{Type} & \textbf{Model} & \textbf{Parameter} & \textbf{FID}$\downarrow$ & \textbf{IS}$\uparrow$ & \textbf{Precision}$\uparrow$ & \textbf{Recall}$\uparrow$ & \textbf{Step} \\
\midrule
\textit{GAN} & BigGAN\,\cite{bigGAN} & 112M & 6.95 & 224.5 & 0.89 & 0.38 & 1\\
\textit{GAN} & GigaGAN\,\cite{kang2023scaling} & 569M & 3.45 & 225.5 & 0.84 & 0.61 & 1 \\
\textit{GAN} & StyleGAN-XL\,\cite{sauer2022stylegan} & 166M & 2.30 & 265.1 & 0.78 & 0.53 & 1 \\ 
\midrule
\textit{Diffusion} & ADM\,\cite{dhariwal2021diffusion} & 554M & 10.94 & 101.0 & 0.69 & 0.63 & 250 \\
\textit{Diffusion} & CDM\,\cite{ho2022cascaded} & - & 4.88 & 158.7 & - & - & 8100 \\
\textit{Diffusion} & LDM-4\,\cite{rombach2022high} & 400M & 3.60 & 247.7 & - & -  & 250 \\
\textit{Diffusion} & DiT-XL/2\,\cite{peebles2023scalable} & 675M & 2.27 & 278.2 & 0.83 & 0.57 & 250 \\
\midrule
\textit{Masked} AR & MaskGIT\,\cite{chang2022maskgit} & 227M & 6.18 & 182.1 & 0.80 & 0.51 & 8 \\
\textit{Masked} AR  & MaskGIT-re\,\cite{li2023mage} & 227M & 4.02 & 355.6 & - & 8 \\
\textit{Masked} AR  & MAGE\,\cite{li2024autoregressiveimagegenerationvector} & 230M & 6.93 & 195.8 & - & - & 20 \\
\midrule
\textit{Next-token} AR & VQGAN\,\cite{esser2021taming} & 227M & 18.65 & 80.4 & 0.78 & 0.26 & 256 \\
\textit{Next-token} AR & VQGAN \,\cite{esser2021taming} & 1.4B & 15.76 & 74.3 & - & - & 256 \\
\textit{Next-token} AR & VQGAN-re\,\cite{yu2021vector} & 1.4B & 5.20 & 280.3 & - & - & 256 \\
\textit{Next-token} AR & ViT-VQGAN\,\cite{yu2021vector} & 1.7B & 4.17 & 175.1 & - & - & 1024 \\
\textit{Next-token} AR & ViT-VQGAN-re\,\cite{yu2021vector} & 1.7B & 3.04 & 227.4 & - & - & 1024 \\
\textit{Next-token} AR & RQTran\,\cite{lee2022autoregressive} & 3.8B & 7.55 & 80.4 & 0.78 & 0.26 & 68 \\
\textit{Next-token} AR & RQTran-re\,\cite{lee2022autoregressive} & 3.8B & 3.80 & 323.7 & - & - & 68 \\
\textit{Next-token} AR & LlamaGen-B\,\cite{llamagen} & 111M & 5.46 & 193.6 & 0.83 & 0.45 & 256 \\
\textit{Next-token} AR & LlamaGen-L\,\cite{llamagen} & 343M & 3.81 & 248.3 & 0.83 & 0.52 & 256 \\
\textit{Next-token} AR & LlamaGen-XL\,\cite{llamagen} & 775M & 3.39 & 227.1 & 0.81 & 0.54 & 256 \\
\textit{Next-token} AR & LlamaGen-XXL\,\cite{llamagen} & 1.4B & 3.09 & 253.6 & 0.83 & 0.53 & 256 \\
\midrule
\rowcolor{lightblue}
\textit{Markovian Scale AR} & Markov-VAR-\textit{d16} & 329.0M & 3.23 & 256.2 & 0.84 & 0.52 & 10 \\
\rowcolor{lightblue}
\textit{Markovian Scale AR} & Markov-VAR-\textit{d20} & 623.2M & 2.44 & 286.1 & 0.83 & 0.56 & 10 \\
\rowcolor{lightblue}
\textit{Markovian Scale AR} & Markov-VAR-\textit{d24} & 1.02B & 2.15 & 310.9 & 0.83 & 0.59 & 10 \\
\bottomrule
\end{tabular}

\vspace{-0.4cm}
\end{table*}

\vspace{-0.2cm}
\subsection{Main Results}
\paragraph{Generation Performance.} We first evaluate the generation performance of Markov-VAR. We conduct comparsion with VAR and other VAR-like models~\cite{han2025infinity,ren2024mvardecoupledscalewiseautoregressive,kumbong2025hmar,wu2025nestedautoregressivemodels} on ImageNet $256 \times 256$ class-conditional image generation. As shown in Table~\ref{tab:var_like}, Markov-VAR demonstrates a comprehensively better performance, achieving higher generation quality than other VAR models of comparable size. For example, on the \textit{depth-16} model, Markov-VAR improve the FID performance from 3.61 to 3.23 (10.5\%) and IS performance from 225.6 to 256.2 (13.6\%) compared to VAR. Even on Precision and Recall metrics, Markov-VAR matches or surpasses VAR. Markov-VAR's advantages can be observed across other sizes. Compared with M-VAR-\textit{d20}, Markov-VAR-\textit{d20} achieves compelling performance while using only 70\% of its parameters, demonstrating that Markov-VAR is both effective and parameter-efficient.

To further demonstrate the performance advancements of Markov-VAR. We conduct a board comparison on class-to-image generation on ImageNet $256 \times 256$ benchmark with more alternative-paradigm models, including GANs~\cite{bigGAN,kang2023scaling,sauer2022stylegan}, diffusion models~\cite{dhariwal2021diffusion,ho2020denoising,rombach2022high,peebles2023scalable}, mask autoregressive models~\cite{chang2022maskgit,li2023mage,li2024autoregressiveimagegenerationvector}, and next-token prediction autoregressive models~\cite{llamagen,esser2021taming,yu2021vector,lee2022autoregressive}. As shown in Table~\ref{tab:board}, Compared with Diffusion models, Markov-VAR achieves better overall performance in terms of both parameter efficiency and generation quality. Compared to the size-similar next-token autoregressive model LlamaGen-L, Markov-VAR achieves superior performance across all evaluation metrics. In contrast to other autoregressive models, Markov-VAR maintains competitive generation quality while requiring fewer sampling steps, demonstrating its higher inference efficiency. Further comparisons in the table also reveal Markov-VAR's favorable generation performance and parameter efficiency.

\paragraph{Generation Efficiency.}
We also evaluate the computational efficiency of Markov-VAR with various resolutions and depths compared to VAR~\cite{VAR} and other VAR-like models~\cite{kumbong2025hmar, jiao2025flexvarflexiblevisualautoregressive}. As shown in Table~\ref{tab:efficiency}, Markov-VAR achieve consistently excellent performance on efficiency metrics. For peak memory consumption, at $256 \times 256$, Markov-VAR-$d16$ reduces it by 56.9\%, and Markov-VAR-$d24$ reduces it by 62.1\%. As resolution increases, the memory advantage becomes increasingly evident: Markov-VAR-$d24$ brings peak memory consumption down from 117.9GB to 19.1GB (83.8\%) at $1024 \times 1024$. Moreover, compared to the other VAR-like model FlexVAR, Markov-VAR achieves an acceleration of 1.33$\times$ for inference at $256 \times 256$. These results demonstrate the efficiency of Markov-VAR.

\begin{table}[h]
    \centering
    \caption{Comparison of inference time and peak memory between Markov-VAR and other VAR-like models across various resolutions and depths. Peak GPU memory is reported as the average of 5 runs with a batch size of 25 on a single NVIDIA H200. Inference time is measured in seconds per batch on a NVIDIA H200.}
    \label{tab:efficiency}
    \vspace{-0.2cm}
    \resizebox{0.48\textwidth}{!}{
        \begin{tabular}{c|cc|cc}
            \toprule
            \textbf{Model} & \textbf{Depth} & \textbf{Res.} & \textbf{Time(s)} $\downarrow$ & \textbf{Memory(GB)} $\downarrow$ \\
            \midrule
            VAR~\cite{VAR}        & 16 & 256  & 0.303 & 5.8 \\
            VAR~\cite{VAR}        & 16 & 512  & 0.824 & 18.4 \\
            VAR~\cite{VAR}        & 16 & 1024 & 3.303 & 66.7 \\
            HMAR~\cite{kumbong2025hmar} & 16 & 256 & 0.309 & 3.1\\
            HMAR~\cite{kumbong2025hmar} & 16 & 512 & 0.909 & 5.6 \\
            FlexVAR~\cite{jiao2025flexvarflexiblevisualautoregressive} & 16 & 256 & 0.395 & 6.8 \\
            FlexVAR~\cite{jiao2025flexvarflexiblevisualautoregressive} & 16 & 512 & 0.944 & 20.1 \\
            FlexVAR~\cite{jiao2025flexvarflexiblevisualautoregressive} & 16 & 1024 & 3.409 & 77.9 \\
            \rowcolor{lightblue}
            Markov-VAR & 16 & 256  & 0.296 & 2.5 \\
            \rowcolor{lightblue}
            Markov-VAR & 16 & 512  & 0.780 & 4.9 \\
            \rowcolor{lightblue}
            Markov-VAR & 16 & 1024 & 3.125 & 14.6 \\
            \midrule
            VAR~\cite{VAR}        & 24 & 256  & 0.711 & 12.4 \\
            VAR~\cite{VAR}        & 24 & 512  & 1.335 & 31.4 \\
            VAR~\cite{VAR}        & 24 & 1024 & 5.891 & 117.9 \\
            \rowcolor{lightblue}
            Markov-VAR & 24 & 256  & 0.608 & 4.7 \\
            \rowcolor{lightblue}
            Markov-VAR & 24 & 512  & 1.261 & 8.1 \\
            \rowcolor{lightblue}
            Markov-VAR & 24 & 1024 & 5.322 & 19.1 \\
            \bottomrule
        \end{tabular}
    }
\vspace{-0.4cm}
\end{table}

\subsection{Ablation and Analysis}

\paragraph{Ablation on History Compensation Mechanism.}

To verify the effectiveness of our sliding-window-based history compensation mechanism, we conduct ablation and further compare ours to other mechanisms. Global history is a full-context compensation, which continuously fuses and updates all previous scales, and hybrid history combines both full-context and non-full-context compensation. In Table~\ref{tab:abb:memory}, all types of compensation mechanism improve generation quality compared to the one without history compensation. However, our sliding-window-based history compensation mechanism achieves the overall best preservation of historical information. These validate the effectiveness of our design and are consistent with the analysis in Section~\ref{method}.

\begin{table}[h]
    \vspace{-0.1cm}
    \centering
    \caption{History compensation ablation on the depth-16 model across different history compensation mechanisms.}
    \label{tab:abb:memory}
    \vspace{-0.2cm}
    \begin{tabular}{l|cc|cc}
        \toprule
        \textbf{Method} & \textbf{Depth} & \textbf{Param} & \textbf{FID $\downarrow$} & \textbf{IS $\uparrow$} \\
        \midrule
        w/o History & 16 & 300M & 3.64 & 247.7 \\
        Global History & 16 & 324M & 3.41 & 245.2 \\
        Hybrid History & 16 & 359M & 3.45 & 257.4 \\
        \rowcolor{lightblue}
        Ours & 16 & 329M & 3.23 & 256.2 \\
        \bottomrule
    \end{tabular}
 \vspace{-0.4cm}
\end{table}

\paragraph{Ablation on Sliding Window Size.}
To determine a suitable sliding window size for Markov-VAR's history compensation mechanism, we conduct an ablation study on Markov-VAR with depths of 16 and 20. As shown in Table~\ref{tab:windowsize}, models with the two depths consistently achieve their best performance when window size is set to 3. In comparison, when the window size is set to 1, Markov-VAR-\textit{d16} respectively decreases FID and IS to 3.53 and 237.8. All these results are also consistent with the analysis of cross-scale interference discussed in Introduction that full-context dependency may hinders distinctive scale feature learning, and the most recent three scales typically have a positive effect on learning.

\begin{table}[h]
    \centering
    \caption{Ablation study of the sliding-window-based history compensation mechanism with different window sizes.}
    \vspace{-0.2cm}
    \label{tab:windowsize}
    \resizebox{0.48\textwidth}{!}{
        \begin{tabular}{c|cc|cc}
            \toprule
            \textbf{Window Size} & \textbf{FID(d16)} $\downarrow$ & \textbf{IS(d16)} $\uparrow$ & \textbf{FID(d20)} $\downarrow$ & \textbf{IS(d20)} $\uparrow$ \\
            \midrule
            1 & 3.53 & 237.8 & 2.50 & 267.9 \\
            2 & 3.39 & 248.6 & 2.47 & 281.4 \\
            3 & 3.23 & 256.2 & 2.44 & 286.1 \\
            4 & 3.33 & 252.3 & 2.56 & 278.2 \\
            \bottomrule
        \end{tabular}
    }
    \vspace{-0.5cm}
\end{table}

\begin{figure}
    \centering
    \includegraphics[width=0.48\textwidth]{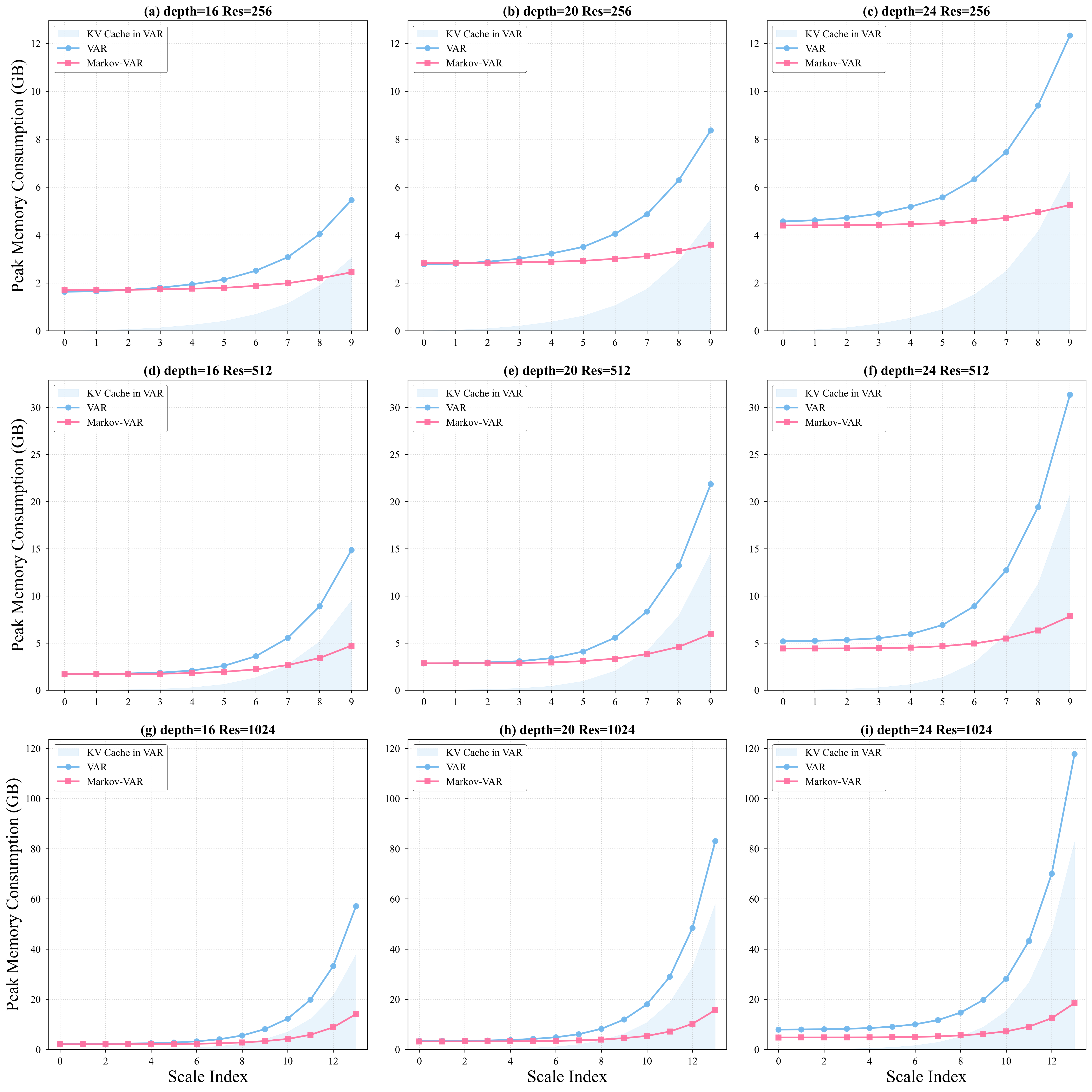}
    \vspace{-0.6cm}
    \caption{Analysis of peak memory consumption of Markov-VAR across various depths and resolutions at different scales.}
    \label{fig:peak_scale}
    \vspace{-0.5cm}
\end{figure}

\paragraph{Memory Consumption Analysis.}
As shown in Figure~\ref{fig:peak_scale}, we investigated the peak memory consumption of computation state across various depths and resolutions at different scales. Compared with VAR, Markov-VAR exhibits a less steep scaling trend rather than the exponential growth trend observed in VAR. As the model size and target resolution increase, the efficiency advantage of Markov-VAR becomes increasingly evident. Most importantly, because Markov-VAR follows a Markovian modeling process, it does not require any KV-cache computation, which fundamentally accounts for its significantly lower computational cost. Moreover, even the computation of Transformer activation in VAR is also larger than that of Markov-VAR, further demonstrating the computational efficiency advantage of Markov-VAR. All these results demonstrate that Markovian Scale Prediction effectively alleviates the exponential growth of full-context modeling, making it more promising for scaling and generating visual content at larger scales.

\begin{figure*}
    \centering
    \includegraphics[width=1\textwidth]{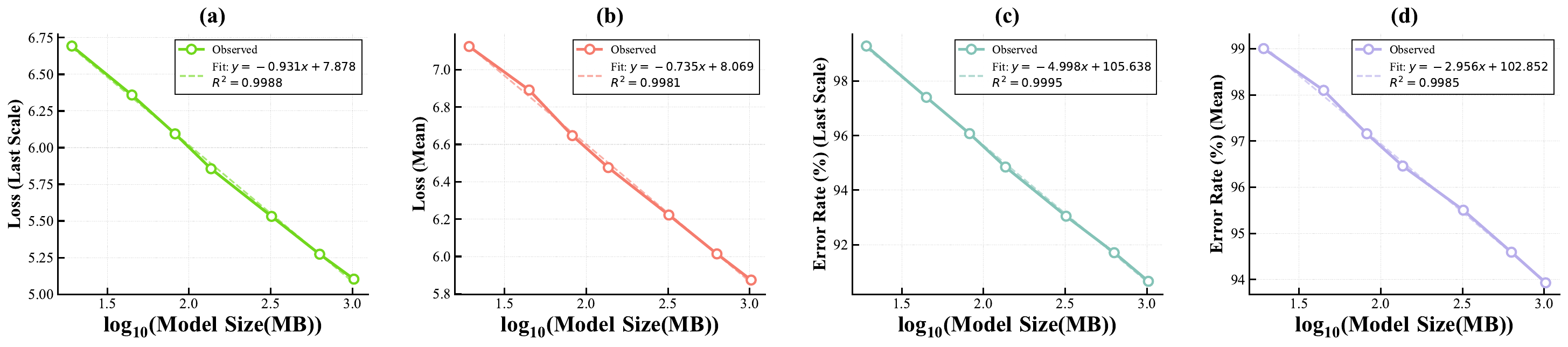}
    \vspace{-0.7cm}
    \caption{Scaling law analysis of Markov-VAR between performance metrics and model sizes with power-law fitted equations.}
    \label{fig:scalinglaw}
    \vspace{-0.1cm}
\end{figure*}

\begin{figure*}
\centering
\includegraphics[width=\linewidth]{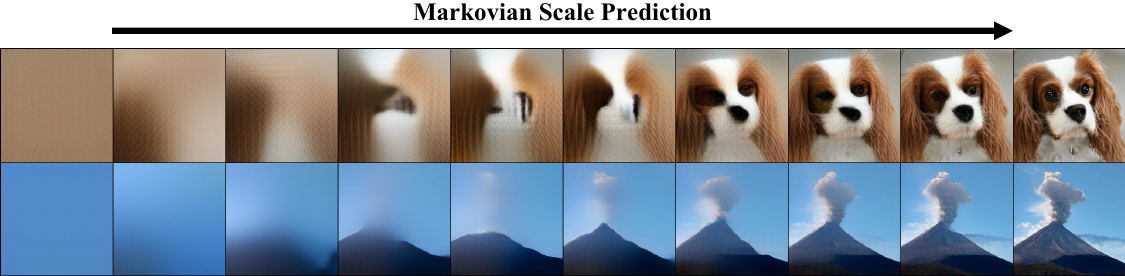}
\vspace{-0.5cm}
\caption{Visualization of generation process in Markov-VAR at $256 \times 256$ resolution.}
\label{fig:scale_level}
\vspace{-0.3cm}
\end{figure*}

\paragraph{Analysis of Scaling Law.}
We vary the depth of Markov-VAR from 6 to 24 to analyze the scaling law, corresponding to parameter scales ranging from 19.80M to 1.02B. In Figure~\ref{fig:scalinglaw}, charts (a) to (d) show the fitted scaling relation between model size (log\textsubscript{10}) and performance metrics. Both loss and error rate consistently decrease as model size increases, following obvious power-law trends with high coefficients of determination ($R^2 > 0.99$), indicating that Markov-VAR follows the expected scaling law.

\subsection{Visualization}
\paragraph{Main Generation Visualization.} Figure~\ref{teaser} shows representative image samples generated by our Markov-VAR. Given the scale and quality of ImageNet, Markov-VAR performs reasonably well. Its generation quality may further improve when trained on larger and higher-quality datasets.

\paragraph{Visualization of Generation Process}
We visualize the Markovian scale prediction process in Figure~\ref{fig:scale_level}. This visualization provides an intuitive understanding of how Markov-VAR constructs a semantic structure and fills in fine-grained details across multiple scales. The earliest scales capture only very coarse global structures.
As the scale increases, the model gradually refines the semantics. Later scales enrich high-frequency textures and refine local details such as fur patterns and cloud shapes. Notably, the refinement is both smooth and consistent, demonstrating that each Markovian scale effectively preserves essential historical information without relying on full-context dependency.

\begin{figure}
    \centering
    \includegraphics[width=\linewidth]{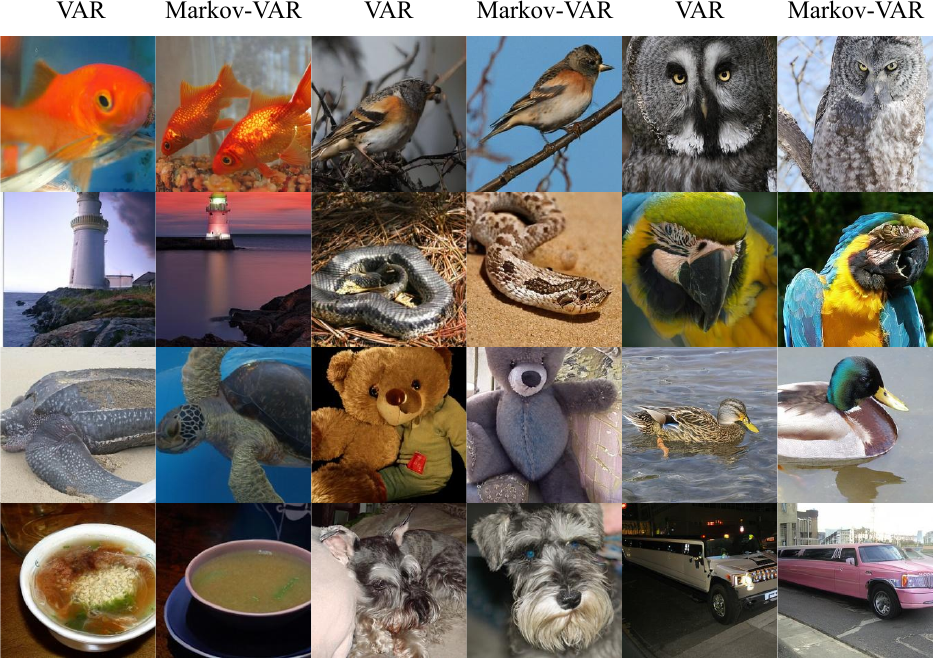}
    \vspace{-0.5cm}
    \caption{Visual comparison between VAR and Markov-VAR.}
    \label{comparison}
    \vspace{-0.4cm}
\end{figure}

\paragraph{Visual Comparison between VAR and Markov-VAR.} Figure~\ref{comparison} shows that both models perform well on certain generated samples, even in some cases, Markov-VAR even produces better semantics and higher quality than VAR.

\section{Conclusion}
In this paper, we address the challenges of VAR's full-context dependency by reformulating it as a Markov process. We propose Markov-VAR, which employs Markovian scale prediction by treating each scale as a Markov state and modeling its evolution as a Markov process. Experiments and analysis demonstrate that Markov-VAR achieves effective and efficient visual generation.

\clearpage
\newpage

{
    \small
    \bibliographystyle{ieeenat_fullname}
    \bibliography{main}

@String(CVPR= {IEEE Conf. Comput. Vis. Pattern Recog.})

@String(CVPR  = {CVPR})

@article{markov1906rasprostranenie,
  title={Rasprostranenie zakona bol’shih chisel na velichiny, zavisyaschie drug ot druga},
  author={Markov, Andrei Andreevich},
  journal={Izvestiya Fiziko-matematicheskogo obschestva pri Kazanskom universitete},
  volume={15},
  number={135-156},
  pages={18},
  year={1906}
}

@article{nie2025large,
  title={Large language diffusion models},
  author={Nie, Shen and Zhu, Fengqi and You, Zebin and Zhang, Xiaolu and Ou, Jingyang and Hu, Jun and Zhou, Jun and Lin, Yankai and Wen, Ji-Rong and Li, Chongxuan},
  journal={arXiv preprint arXiv:2502.09992},
  year={2025}
}

@article{zhu2025llada,
  title={LLaDA 1.5: Variance-Reduced Preference Optimization for Large Language Diffusion Models},
  author={Zhu, Fengqi and Wang, Rongzhen and Nie, Shen and Zhang, Xiaolu and Wu, Chunwei and Hu, Jun and Zhou, Jun and Chen, Jianfei and Lin, Yankai and Wen, Ji-Rong and others},
  journal={arXiv preprint arXiv:2505.19223},
  year={2025}
}

@article{croitoru2023diffusion,
  title={Diffusion models in vision: A survey},
  author={Croitoru, Florinel-Alin and Hondru, Vlad and Ionescu, Radu Tudor and Shah, Mubarak},
  journal={IEEE transactions on pattern analysis and machine intelligence},
  volume={45},
  number={9},
  pages={10850--10869},
  year={2023},
  publisher={Ieee}
}

@article{zhuang2025vargpt,
  title={Vargpt-v1. 1: Improve visual autoregressive large unified model via iterative instruction tuning and reinforcement learning},
  author={Zhuang, Xianwei and Xie, Yuxin and Deng, Yufan and Yang, Dongchao and Liang, Liming and Ru, Jinghan and Yin, Yuguo and Zou, Yuexian},
  journal={arXiv preprint arXiv:2504.02949},
  year={2025}
}

@article{ddpm,
  title   = {Denoising diffusion probabilistic models},
  author  = {Ho, Jonathan and Jain, Ajay and Abbeel, Pieter},
  journal = {Advances in neural information processing systems},
  volume  = {33},
  pages   = {6840--6851},
  year    = {2020}
}

@article{ddim,
  title   = {Denoising diffusion implicit models},
  author  = {Song, Jiaming and Meng, Chenlin and Ermon, Stefano},
  journal = {arXiv preprint arXiv:2010.02502},
  year    = {2020}
}

@article{liu2025marcos,
  title={MARCOS: Deep Thinking by Markov Chain of Continuous Thoughts},
  author={Liu, Jiayu and Huang, Zhenya and Sims, Anya and Chen, Enhong and Teh, Yee Whye and Miao, Ning},
  journal={arXiv preprint arXiv:2509.25020},
  year={2025}
}

@article{yang2024markov,
  title={Markov chain of thought for efficient mathematical reasoning},
  author={Yang, Wen and Liao, Minpeng and Fan, Kai},
  journal={arXiv preprint arXiv:2410.17635},
  year={2024}
}

@article{VAR,
  title={Visual autoregressive modeling: Scalable image generation via next-scale prediction},
  author={Tian, Keyu and Jiang, Yi and Yuan, Zehuan and Peng, Bingyue and Wang, Liwei},
  journal={Advances in neural information processing systems},
  volume={37},
  pages={84839--84865},
  year={2024}
}

@inproceedings{chang2022maskgit,
  title={Maskgit: Masked generative image transformer},
  author={Chang, Huiwen and Zhang, Han and Jiang, Lu and Liu, Ce and Freeman, William T},
  booktitle={Proceedings of the IEEE/CVF conference on computer vision and pattern recognition},
  pages={11315--11325},
  year={2022}
}

@article{lu2025easytext,
  title={EasyText: Controllable Diffusion Transformer for Multilingual Text Rendering},
  author={Lu, Runnan and Zhang, Yuxuan and Liu, Jiaming and Wang, Haofan and Song, Yiren},
  journal={arXiv preprint arXiv:2505.24417},
  year={2025}
}

@inproceedings{kumbong2025hmar,
  title={HMAR: Efficient Hierarchical Masked Auto-Regressive Image Generation},
  author={Kumbong, Hermann and Liu, Xian and Lin, Tsung-Yi and Liu, Ming-Yu and Liu, Xihui and Liu, Ziwei and Fu, Daniel Y and Re, Christopher and Romero, David W},
  booktitle={Proceedings of the Computer Vision and Pattern Recognition Conference},
  pages={2535--2544},
  year={2025}
}

@article{shi2024fonts,
  title={FonTS: Text Rendering with Typography and Style Controls},
  author={Shi, Wenda and Song, Yiren and Zhang, Dengming and Liu, Jiaming and Zou, Xingxing},
  journal={arXiv preprint arXiv:2412.00136},
  year={2024}
}

@article{shi2025wordcon,
  title={WordCon: Word-level Typography Control in Scene Text Rendering},
  author={Shi, Wenda and Song, Yiren and Rao, Zihan and Zhang, Dengming and Liu, Jiaming and Zou, Xingxing},
  journal={arXiv preprint arXiv:2506.21276},
  year={2025}
}

@article{touvron2023llama,
  title={Llama: Open and efficient foundation language models},
  author={Touvron, Hugo and Lavril, Thibaut and Izacard, Gautier and Martinet, Xavier and Lachaux, Marie-Anne and Lacroix, Timoth{\'e}e and Rozi{\`e}re, Baptiste and Goyal, Naman and Hambro, Eric and Azhar, Faisal and others},
  journal={arXiv preprint arXiv:2302.13971},
  year={2023}
}

@article{chung2024scaling,
  title={Scaling instruction-finetuned language models},
  author={Chung, Hyung Won and Hou, Le and Longpre, Shayne and Zoph, Barret and Tay, Yi and Fedus, William and Li, Yunxuan and Wang, Xuezhi and Dehghani, Mostafa and Brahma, Siddhartha and others},
  journal={Journal of Machine Learning Research},
  volume={25},
  number={70},
  pages={1--53},
  year={2024}
}

@article{zhang2025mvar,
  title={MVAR: Visual Autoregressive Modeling with Scale and Spatial Markovian Conditioning},
  author={Zhang, Jinhua and Long, Wei and Han, Minghao and You, Weiyi and Gu, Shuhang},
  journal={arXiv preprint arXiv:2505.12742},
  year={2025}
}

@article{hui2025autoregressive,
  title={Autoregressive Images Watermarking through Lexical Biasing: An Approach Resistant to Regeneration Attack},
  author={Hui, Siqi and Song, Yiren and Zhou, Sanping and Deng, Ye and Huang, Wenli and Wang, Jinjun},
  journal={arXiv preprint arXiv:2506.01011},
  year={2025}
}

@article{wan2024grid,
  title={Grid: Visual layout generation},
  author={Wan, Cong and Luo, Xiangyang and Cai, Zijian and Song, Yiren and Zhao, Yunlong and Bai, Yifan and He, Yuhang and Gong, Yihong},
  journal={arXiv preprint arXiv:2412.10718},
  year={2024}
}

@article{song2025makeanything,
  title={MakeAnything: Harnessing Diffusion Transformers for Multi-Domain Procedural Sequence Generation},
  author={Song, Yiren and Liu, Cheng and Shou, Mike Zheng},
  journal={arXiv preprint arXiv:2502.01572},
  year={2025}
}

@article{zhang2025easycontrol,
  title={Easycontrol: Adding efficient and flexible control for diffusion transformer},
  author={Zhang, Yuxuan and Yuan, Yirui and Song, Yiren and Wang, Haofan and Liu, Jiaming},
  journal={arXiv preprint arXiv:2503.07027},
  year={2025}
}

@article{achiam2023gpt,
  title={Gpt-4 technical report},
  author={Achiam, Josh and Adler, Steven and Agarwal, Sandhini and Ahmad, Lama and Akkaya, Ilge and Aleman, Florencia Leoni and Almeida, Diogo and Altenschmidt, Janko and Altman, Sam and Anadkat, Shyamal and others},
  journal={arXiv preprint arXiv:2303.08774},
  year={2023}
}

@inproceedings{esser2021taming,
  title={Taming transformers for high-resolution image synthesis},
  author={Esser, Patrick and Rombach, Robin and Ommer, Bjorn},
  booktitle={Proceedings of the IEEE/CVF conference on computer vision and pattern recognition},
  pages={12873--12883},
  year={2021}
}

@article{chen2025transanimate,
  title={Transanimate: Taming layer diffusion to generate rgba video},
  author={Chen, Xuewei and Chen, Zhimin and Song, Yiren},
  journal={arXiv preprint arXiv:2503.17934},
  year={2025}
}

@article{llamagen,
  title={Autoregressive Model Beats Diffusion: Llama for Scalable Image Generation},
  author={Sun, Peize and Jiang, Yi and Chen, Shoufa and Zhang, Shilong and Peng, Bingyue and Luo, Ping and Yuan, Zehuan},
  journal={arXiv preprint arXiv:2406.06525},
  year={2024}
}

@article{van2017neural,
  title={Neural discrete representation learning},
  author={Van Den Oord, Aaron and Vinyals, Oriol and others},
  journal={Advances in neural information processing systems},
  volume={30},
  year={2017}
}

@article{van2016conditional,
  title={Conditional image generation with pixelcnn decoders},
  author={Van den Oord, Aaron and Kalchbrenner, Nal and Espeholt, Lasse and Vinyals, Oriol and Graves, Alex and others},
  journal={Advances in neural information processing systems},
  volume={29},
  year={2016}
}

@article{esser2021imagebart,
  title={Imagebart: Bidirectional context with multinomial diffusion for autoregressive image synthesis},
  author={Esser, Patrick and Rombach, Robin and Blattmann, Andreas and Ommer, Bjorn},
  journal={Advances in neural information processing systems},
  volume={34},
  pages={3518--3532},
  year={2021}
}

@article{zhu2024scaling,
  title={Scaling the codebook size of VQ-GAN to 100,000 with a utilization rate of 99\%},
  author={Zhu, Lei and Wei, Fangyun and Lu, Yanye and Chen, Dong},
  journal={Advances in Neural Information Processing Systems},
  volume={37},
  pages={12612--12635},
  year={2024}
}

@inproceedings{rqvae,
  title={Autoregressive image generation using residual quantization},
  author={Lee, Doyup and Kim, Chiheon and Kim, Saehoon and Cho, Minsu and Han, Wook-Shin},
  booktitle={Proceedings of the IEEE/CVF conference on computer vision and pattern recognition},
  pages={11523--11532},
  year={2022}
}

@article{zhang2024var,
  title={Var-clip: Text-to-image generator with visual auto-regressive modeling},
  author={Zhang, Qian and Dai, Xiangzi and Yang, Ninghua and An, Xiang and Feng, Ziyong and Ren, Xingyu},
  journal={arXiv preprint arXiv:2408.01181},
  year={2024}
}

@inproceedings{han2025infinity,
  title={Infinity: Scaling bitwise autoregressive modeling for high-resolution image synthesis},
  author={Han, Jian and Liu, Jinlai and Jiang, Yi and Yan, Bin and Zhang, Yuqi and Yuan, Zehuan and Peng, Bingyue and Liu, Xiaobing},
  booktitle={Proceedings of the Computer Vision and Pattern Recognition Conference},
  pages={15733--15744},
  year={2025}
}

@article{salimans2017pixelcnn++,
  title={Pixelcnn++: Improving the pixelcnn with discretized logistic mixture likelihood and other modifications},
  author={Salimans, Tim and Karpathy, Andrej and Chen, Xi and Kingma, Diederik P},
  journal={arXiv preprint arXiv:1701.05517},
  year={2017}
}

@article{tang2024hart,
  title={Hart: Efficient visual generation with hybrid autoregressive transformer},
  author={Tang, Haotian and Wu, Yecheng and Yang, Shang and Xie, Enze and Chen, Junsong and Chen, Junyu and Zhang, Zhuoyang and Cai, Han and Lu, Yao and Han, Song},
  journal={arXiv preprint arXiv:2410.10812},
  year={2024}
}

@inproceedings{kumari2023multi,
  title={Multi-concept customization of text-to-image diffusion},
  author={Kumari, Nupur and Zhang, Bingliang and Zhang, Richard and Shechtman, Eli and Zhu, Jun-Yan},
  booktitle={Proceedings of the IEEE/CVF conference on computer vision and pattern recognition},
  pages={1931--1941},
  year={2023}
}

@article{goodfellow2014generative,
  title={Generative adversarial nets},
  author={Goodfellow, Ian J and Pouget-Abadie, Jean and Mirza, Mehdi and Xu, Bing and Warde-Farley, David and Ozair, Sherjil and Courville, Aaron and Bengio, Yoshua},
  journal={Advances in neural information processing systems},
  volume={27},
  year={2014}
}

@article{wang2025training,
  title={Training-Free Text-Guided Image Editing with Visual Autoregressive Model},
  author={Wang, Yufei and Guo, Lanqing and Li, Zhihao and Huang, Jiaxing and Wang, Pichao and Wen, Bihan and Wang, Jian},
  journal={arXiv preprint arXiv:2503.23897},
  year={2025}
}

@misc{su2023roformerenhancedtransformerrotary,
      title={RoFormer: Enhanced Transformer with Rotary Position Embedding}, 
      author={Jianlin Su and Yu Lu and Shengfeng Pan and Ahmed Murtadha and Bo Wen and Yunfeng Liu},
      year={2023},
      eprint={2104.09864},
      archivePrefix={arXiv},
      primaryClass={cs.CL},
      url={https://arxiv.org/abs/2104.09864}, 
}

@misc{wu2025nestedautoregressivemodels,
      title={Nested AutoRegressive Models}, 
      author={Hongyu Wu and Xuhui Fan and Zhangkai Wu and Longbing Cao},
      year={2025},
      eprint={2510.23028},
      archivePrefix={arXiv},
      primaryClass={cs.CV},
      url={https://arxiv.org/abs/2510.23028}, 
}

@article{ren2024mvardecoupledscalewiseautoregressive,
      title={M-VAR: Decoupled Scale-wise Autoregressive Modeling for High-Quality Image Generation}, 
      author={Sucheng Ren and Yaodong Yu and Nataniel Ruiz and Feng Wang and Alan Yuille and Cihang Xie},
      year={2024},
      eprint={2411.10433},
      archivePrefix={arXiv},
      primaryClass={cs.CV},
      url={https://arxiv.org/abs/2411.10433}, 
}

@article{jiao2025flexvarflexiblevisualautoregressive,
  title={Flexvar: Flexible visual autoregressive modeling without residual prediction},
  author={Jiao, Siyu and Zhang, Gengwei and Qian, Yinlong and Huang, Jiancheng and Zhao, Yao and Shi, Humphrey and Ma, Lin and Wei, Yunchao and Jie, Zequn},
  journal={arXiv preprint arXiv:2502.20313},
  year={2025}
}

@article{bigGAN,
  title={Large scale GAN training for high fidelity natural image synthesis},
  author={Brock, Andrew and Donahue, Jeff and Simonyan, Karen},
  journal={arXiv preprint arXiv:1809.11096},
  year={2018}
}

@inproceedings{kang2023scaling,
  title={Scaling up gans for text-to-image synthesis},
  author={Kang, Minguk and Zhu, Jun-Yan and Zhang, Richard and Park, Jaesik and Shechtman, Eli and Paris, Sylvain and Park, Taesung},
  booktitle={Proceedings of the IEEE/CVF conference on computer vision and pattern recognition},
  pages={10124--10134},
  year={2023}
}

@inproceedings{sauer2022stylegan,
  title={Stylegan-xl: Scaling stylegan to large diverse datasets},
  author={Sauer, Axel and Schwarz, Katja and Geiger, Andreas},
  booktitle={ACM SIGGRAPH 2022 conference proceedings},
  pages={1--10},
  year={2022}
}

@article{dhariwal2021diffusion,
  title={Diffusion models beat gans on image synthesis},
  author={Dhariwal, Prafulla and Nichol, Alexander},
  journal={Advances in neural information processing systems},
  volume={34},
  pages={8780--8794},
  year={2021}
}

@article{ho2022cascaded,
  title={Cascaded diffusion models for high fidelity image generation},
  author={Ho, Jonathan and Saharia, Chitwan and Chan, William and Fleet, David J and Norouzi, Mohammad and Salimans, Tim},
  journal={Journal of Machine Learning Research},
  volume={23},
  number={47},
  pages={1--33},
  year={2022}
}

@inproceedings{rombach2022high,
  title={High-resolution image synthesis with latent diffusion models},
  author={Rombach, Robin and Blattmann, Andreas and Lorenz, Dominik and Esser, Patrick and Ommer, Bj{\"o}rn},
  booktitle={Proceedings of the IEEE/CVF conference on computer vision and pattern recognition},
  pages={10684--10695},
  year={2022}
}

@inproceedings{peebles2023scalable,
  title={Scalable diffusion models with transformers},
  author={Peebles, William and Xie, Saining},
  booktitle={Proceedings of the IEEE/CVF international conference on computer vision},
  pages={4195--4205},
  year={2023}
}

@inproceedings{li2023mage,
  title={Mage: Masked generative encoder to unify representation learning and image synthesis},
  author={Li, Tianhong and Chang, Huiwen and Mishra, Shlok and Zhang, Han and Katabi, Dina and Krishnan, Dilip},
  booktitle={Proceedings of the IEEE/CVF Conference on Computer Vision and Pattern Recognition},
  pages={2142--2152},
  year={2023}
}

@inproceedings{chen2025sar3d,
  title={SAR3D: Autoregressive 3D object generation and understanding via multi-scale 3D VQVAE},
  author={Chen, Yongwei and Lan, Yushi and Zhou, Shangchen and Wang, Tengfei and Pan, Xingang},
  booktitle={Proceedings of the Computer Vision and Pattern Recognition Conference},
  pages={28371--28382},
  year={2025}
}

@misc{li2024autoregressiveimagegenerationvector,
  title         = {Autoregressive Image Generation without Vector Quantization},
  author        = {Tianhong Li and Yonglong Tian and He Li and Mingyang Deng and Kaiming He},
  year          = {2024},
  eprint        = {2406.11838},
  archiveprefix = {arXiv},
  primaryclass  = {cs.CV},
  url           = {https://arxiv.org/abs/2406.11838}
}

@article{yu2021vector,
  title={Vector-quantized image modeling with improved vqgan},
  author={Yu, Jiahui and Li, Xin and Koh, Jing Yu and Zhang, Han and Pang, Ruoming and Qin, James and Ku, Alexander and Xu, Yuanzhong and Baldridge, Jason and Wu, Yonghui},
  journal={arXiv preprint arXiv:2110.04627},
  year={2021}
}

@inproceedings{lee2022autoregressive,
  title={Autoregressive image generation using residual quantization},
  author={Lee, Doyup and Kim, Chiheon and Kim, Saehoon and Cho, Minsu and Han, Wook-Shin},
  booktitle={Proceedings of the IEEE/CVF Conference on Computer Vision and Pattern Recognition},
  pages={11523--11532},
  year={2022}
}

@article{loshchilov2017decoupled,
  title={Decoupled weight decay regularization},
  author={Loshchilov, Ilya and Hutter, Frank},
  journal={arXiv preprint arXiv:1711.05101},
  year={2017}
}

@article{qu2025visual,
  title={Visual autoregressive modeling for image super-resolution},
  author={Qu, Yunpeng and Yuan, Kun and Hao, Jinhua and Zhao, Kai and Xie, Qizhi and Sun, Ming and Zhou, Chao},
  journal={arXiv preprint arXiv:2501.18993},
  year={2025}
}

@inproceedings{guo2024i2v,
  title={I2v-adapter: A general image-to-video adapter for diffusion models},
  author={Guo, Xun and Zheng, Mingwu and Hou, Liang and Gao, Yuan and Deng, Yufan and Wan, Pengfei and Zhang, Di and Liu, Yufan and Hu, Weiming and Zha, Zhengjun and others},
  booktitle={ACM SIGGRAPH 2024 Conference Papers},
  pages={1--12},
  year={2024}
}

@article{guo2025fastvar,
  title={Fastvar: Linear visual autoregressive modeling via cached token pruning},
  author={Guo, Hang and Li, Yawei and Zhang, Taolin and Wang, Jiangshan and Dai, Tao and Xia, Shu-Tao and Benini, Luca},
  journal={arXiv preprint arXiv:2503.23367},
  year={2025}
}

@article{rajagopalan2025restorevar,
  title={RestoreVAR: Visual Autoregressive Generation for All-in-One Image Restoration},
  author={Rajagopalan, Sudarshan and Narayan, Kartik and Patel, Vishal M},
  journal={arXiv preprint arXiv:2505.18047},
  year={2025}
}

@article{xie2024litevar,
  title={Litevar: Compressing visual autoregressive modelling with efficient attention and quantization},
  author={Xie, Rui and Zhao, Tianchen and Yuan, Zhihang and Wan, Rui and Gao, Wenxi and Zhu, Zhenhua and Ning, Xuefei and Wang, Yu},
  journal={arXiv preprint arXiv:2411.17178},
  year={2024}
}

@article{ho2020denoising,
  title={Denoising diffusion probabilistic models},
  author={Ho, Jonathan and Jain, Ajay and Abbeel, Pieter},
  journal={Advances in neural information processing systems},
  volume={33},
  pages={6840--6851},
  year={2020}
}

@INPROCEEDINGS{7133169,
  author={Tishby, Naftali and Zaslavsky, Noga},
  booktitle={2015 IEEE Information Theory Workshop (ITW)}, 
  title={Deep learning and the information bottleneck principle}, 
  year={2015},
  volume={},
  number={},
  pages={1-5},
  doi={10.1109/ITW.2015.7133169}}

@misc{tishby2000informationbottleneckmethod,
      title={The information bottleneck method}, 
      author={Naftali Tishby and Fernando C. Pereira and William Bialek},
      year={2000},
      eprint={physics/0004057},
      archivePrefix={arXiv},
      primaryClass={physics.data-an},
      url={https://arxiv.org/abs/physics/0004057}, 
}

@misc{alemi2019deepvariationalinformationbottleneck,
      title={Deep Variational Information Bottleneck}, 
      author={Alexander A. Alemi and Ian Fischer and Joshua V. Dillon and Kevin Murphy},
      year={2019},
      eprint={1612.00410},
      archivePrefix={arXiv},
      primaryClass={cs.LG},
      url={https://arxiv.org/abs/1612.00410}, 
}

@inproceedings{xie2025collaboration,
  title={Collaboration Wins More: Dual-Modal Collaborative Attention Reinforcement for Mitigating Large Vision Language Models Hallucination},
  author={Xie, Jiye and Gao, Yifei and You, Liangliang and Xu, Xiang and Xu, Haoran and Kou, Zhiqiang and Fu, Kexue and Qu, Youyang and Yang, Wenjie and Guo, Jianwei and others},
  booktitle={Proceedings of the 33rd ACM International Conference on Multimedia},
  pages={4137--4146},
  year={2025}
}

@article{wan2024d2o,
  title={D2o: Dynamic discriminative operations for efficient generative inference of large language models},
  author={Wan, Zhongwei and Wu, Xinjian and Zhang, Yu and Xin, Yi and Tao, Chaofan and Zhu, Zhihong and Wang, Xin and Luo, Siqi and Xiong, Jing and Zhang, Mi},
  journal={arXiv preprint arXiv:2406.13035},
  year={2024}
}

@article{wan2023efficient,
  title={Efficient large language models: A survey},
  author={Wan, Zhongwei and Wang, Xin and Liu, Che and Alam, Samiul and Zheng, Yu and Liu, Jiachen and Qu, Zhongnan and Yan, Shen and Zhu, Yi and Zhang, Quanlu and others},
  journal={arXiv preprint arXiv:2312.03863},
  year={2023}
}

@inproceedings{deng2009imagenet,
  title={Imagenet: A large-scale hierarchical image database},
  author={Deng, Jia and Dong, Wei and Socher, Richard and Li, Li-Jia and Li, Kai and Fei-Fei, Li},
  booktitle={2009 IEEE conference on computer vision and pattern recognition},
  pages={248--255},
  year={2009},
  organization={Ieee}
}

@inproceedings{van2016pixel,
  title={Pixel recurrent neural networks},
  author={Van Den Oord, A{\"a}ron and Kalchbrenner, Nal and Kavukcuoglu, Koray},
  booktitle={International conference on machine learning},
  pages={1747--1756},
  year={2016},
  organization={PMLR}
}

@inproceedings{ramesh2021zero,
  title={Zero-shot text-to-image generation},
  author={Ramesh, Aditya and Pavlov, Mikhail and Goh, Gabriel and Gray, Scott and Voss, Chelsea and Radford, Alec and Chen, Mark and Sutskever, Ilya},
  booktitle={International conference on machine learning},
  pages={8821--8831},
  year={2021},
  organization={Pmlr}
}

@misc{ding2021cogviewmasteringtexttoimagegeneration,
      title={CogView: Mastering Text-to-Image Generation via Transformers}, 
      author={Ming Ding and Zhuoyi Yang and Wenyi Hong and Wendi Zheng and Chang Zhou and Da Yin and Junyang Lin and Xu Zou and Zhou Shao and Hongxia Yang and Jie Tang},
      year={2021},
      eprint={2105.13290},
      archivePrefix={arXiv},
      primaryClass={cs.CV},
      url={https://arxiv.org/abs/2105.13290}, 
}

@article{li2024controlvar,
  title={Controlvar: Exploring controllable visual autoregressive modeling},
  author={Li, Xiang and Qiu, Kai and Chen, Hao and Kuen, Jason and Lin, Zhe and Singh, Rita and Raj, Bhiksha},
  journal={arXiv preprint arXiv:2406.09750},
  year={2024}
}

@article{shao2025continuous,
  title={Continuous Visual Autoregressive Generation via Score Maximization},
  author={Shao, Chenze and Meng, Fandong and Zhou, Jie},
  journal={arXiv preprint arXiv:2505.07812},
  year={2025}
}

@InProceedings{Chen_2024_CVPR,
    author    = {Chen, Zhe and Wu, Jiannan and Wang, Wenhai and Su, Weijie and Chen, Guo and Xing, Sen and Zhong, Muyan and Zhang, Qinglong and Zhu, Xizhou and Lu, Lewei and Li, Bin and Luo, Ping and Lu, Tong and Qiao, Yu and Dai, Jifeng},
    title     = {InternVL: Scaling up Vision Foundation Models and Aligning for Generic Visual-Linguistic Tasks},
    booktitle = {Proceedings of the IEEE/CVF Conference on Computer Vision and Pattern Recognition (CVPR)},
    month     = {June},
    year      = {2024},
    pages     = {24185-24198}
}

@article{li2024llava,
  title={Llava-next-interleave: Tackling multi-image, video, and 3d in large multimodal models},
  author={Li, Feng and Zhang, Renrui and Zhang, Hao and Zhang, Yuanhan and Li, Bo and Li, Wei and Ma, Zejun and Li, Chunyuan},
  journal={arXiv preprint arXiv:2407.07895},
  year={2024}
}

@misc{yu2022scalingautoregressivemodelscontentrich,
      title={Scaling Autoregressive Models for Content-Rich Text-to-Image Generation}, 
      author={Jiahui Yu and Yuanzhong Xu and Jing Yu Koh and Thang Luong and Gunjan Baid and Zirui Wang and Vijay Vasudevan and Alexander Ku and Yinfei Yang and Burcu Karagol Ayan and Ben Hutchinson and Wei Han and Zarana Parekh and Xin Li and Han Zhang and Jason Baldridge and Yonghui Wu},
      year={2022},
      eprint={2206.10789},
      archivePrefix={arXiv},
      primaryClass={cs.CV},
      url={https://arxiv.org/abs/2206.10789}, 
}

@article{shen2025efficient,
  title={Efficient diffusion models: A survey},
  author={Shen, Hui and Zhang, Jingxuan and Xiong, Boning and Hu, Rui and Chen, Shoufa and Wan, Zhongwei and Wang, Xin and Zhang, Yu and Gong, Zixuan and Bao, Guangyin and others},
  journal={arXiv preprint arXiv:2502.06805},
  year={2025}
}

@inproceedings{zhang2025enhancing,
  title={Enhancing text-to-image diffusion transformer via split-text conditioning},
  author={Zhang, Yu and Zhou, Jialei and Li, Xinchen and Zhang, Qi and Wan, Zhongwei and Miao, Duoqian and Wang, Changwei and Cao, Longbing},
  booktitle={The Thirty-ninth Annual Conference on Neural Information Processing Systems}
}

@article{shannon1948mathematical,
  title={A mathematical theory of communication},
  author={Shannon, Claude E},
  journal={The Bell system technical journal},
  volume={27},
  number={3},
  pages={379--423},
  year={1948},
  publisher={Nokia Bell Labs}
}

@misc{InfinityStar,
      title={InfinityStar: Unified Spacetime AutoRegressive Modeling for Visual Generation}, 
      author={Jinlai Liu and Jian Han and Bin Yan and Hui Wu and Fengda Zhu and Xing Wang and Yi Jiang and Bingyue Peng and Zehuan Yuan},
      year={2025},
      eprint={2511.04675},
      archivePrefix={arXiv},
      primaryClass={cs.CV},
      url={https://arxiv.org/abs/2511.04675}, 
}

@inproceedings{wiewiora2005learning,
  title={Learning predictive representations from a history},
  author={Wiewiora, Eric},
  booktitle={Proceedings of the 22nd international conference on Machine learning},
  pages={964--971},
  year={2005}
}

@inproceedings{yang2022discrete,
  title={Discrete approximate information states in partially observable environments},
  author={Yang, Lujie and Zhang, Kaiqing and Amice, Alexandre and Li, Yunzhu and Tedrake, Russ},
  booktitle={2022 American Control Conference (ACC)},
  pages={1406--1413},
  year={2022},
  organization={IEEE}
}

@inproceedings{mao2020information,
  title={Information state embedding in partially observable cooperative multi-agent reinforcement learning},
  author={Mao, Weichao and Zhang, Kaiqing and Miehling, Erik and Ba{\c{s}}ar, Tamer},
  booktitle={2020 59th IEEE Conference on Decision and Control (CDC)},
  pages={6124--6131},
  year={2020},
  organization={IEEE}
}
}

\end{document}